\documentclass[11pt, a4paper]{article}

\usepackage[margin=1.2in]{geometry} 
\usepackage{newtxtext,newtxmath}    % Times New Roman style font
\usepackage[T1]{fontenc}
\usepackage[utf8]{inputenc}
\usepackage{microtype}              % Improves text justification

\usepackage[numbers,sort&compress]{natbib} 

\usepackage{xurl}   

\usepackage{graphicx}
\usepackage{booktabs}               
\usepackage{caption}
\usepackage{subcaption}             
\usepackage[table,xcdraw]{xcolor}   
\usepackage{float}

\usepackage[colorlinks=true,
            linkcolor=blue!60!black,
            citecolor=blue!60!black,
            urlcolor=blue!60!black,
            breaklinks=true]{hyperref}

\usepackage{algorithm}
\usepackage{algorithmic}

\renewenvironment{abstract}
 {\small
  \begin{center}
  \bfseries \abstractname\vspace{-.5em}\vspace{0pt}
  \end{center}
  \list{}{
    \setlength{\leftmargin}{1cm}
    \setlength{\rightmargin}{1cm}
  }%
  \item\relax}
 {\endlist}

\title{\textbf{\Large GradID: Adversarial Detection via Intrinsic Dimensionality of Gradients}}

\author{
    \normalsize \textbf{Mohammad Mahdi Razmjoo}\textsuperscript{1} \quad 
    \textbf{Mohammad Mahdi Sharifian}\textsuperscript{1} \quad 
    \textbf{Saeed Bagheri Shouraki}\textsuperscript{1} \\[1em]
    \small \textsuperscript{1}Sharif University of Technology \\
    \small \texttt{\{mm.razmjoo, mm.sharifian\}@ee.sharif.edu, bagheri-s@sharif.edu}
}

\date{}

\begin{document}

\maketitle

\begin{abstract}
\noindent 
Despite their remarkable performance, deep neural networks exhibit a critical vulnerability: small, often imperceptible, adversarial perturbations can lead to drastically altered model predictions. Given the stringent reliability demands of applications such as medical diagnosis and autonomous driving, robust detection of such adversarial attacks is paramount.
In this paper, we investigate the geometric properties of a model’s input loss landscape. We analyze the Intrinsic Dimensionality (ID) of the model's gradient parameters, which quantifies the minimal number of coordinates required to describe the data points on their underlying manifold. We reveal a distinct and consistent difference in the ID for natural and adversarial data, which forms the basis of our proposed detection method.
We validate our approach across two distinct operational scenarios. First, in a batch-wise context for identifying malicious data groups, our method demonstrates high efficacy on datasets like MNIST and SVHN. Second, in the critical individual-sample setting, we establish new state-of-the-art results on challenging benchmarks such as CIFAR-10 and MS COCO. Our detector significantly surpasses existing methods against a wide array of attacks, including CW and AutoAttack, achieving detection rates consistently above 92\% on CIFAR-10. The results underscore the robustness of our geometric approach, highlighting that intrinsic dimensionality is a powerful fingerprint for adversarial detection across diverse datasets and attack strategies.
\end{abstract}

% ----------------------------------------------------------------

\section{Introduction}

Deep neural networks have been proven to be vulnerable to adversarial attacks \cite{szegedy2013intriguing, goodfellow2014explaining}. Such attacks intentionally modify sample data to fool the model into making incorrect predictions. This vulnerability raises serious concerns about the safety of such models, especially when used in applications such as medical diagnosis or autonomous driving, which require extensive robustness \cite{eykholt2018robust}. Consequently, mitigating the impact of adversarial attacks has become a topic of extensive research. One general method to address this issue is to distinguish and discard adversarial samples. While the distinguish-and-discard method can be integrated with previous defense methods, reliably differentiating adversarial input data from natural data remains a challenge.

Adversarial detection methods can be categorized based on the criterion of adversariality into two main groups: distribution-based and perturbation-based methods. Distribution-based methods like DDAD \cite{cui2023ddad} and SADD \cite{zhang2022sadd} distinguish adversarial data by considering the statistical distribution of input data samples. These methods possess a rich theoretical basis and can even denoise adversarial data samples rather than merely discarding them, while still maintaining high accuracy rates. However, due to the requirement of processing data in batches, these models incur high computational costs. Perturbation-based models, on the other hand, relate adversariality to the intrinsic properties of the data and attempt to identify adversarial input data based on the perturbation caused by added noise or abnormality \cite{feinman2017detecting}. Such methods define a perturbation metric and distinguish adversarial samples based on that metric. These methods do not require processing data in batch format and therefore have lower computational costs; however, their validation is primarily empirical rather than theoretical. Additionally, selecting an optimal perturbation metric remains an open problem.

The input loss landscape explains the behavior of input data with respect to the loss function. By analyzing the loss surface for a natural and an adversarial data sample, distinct behaviors emerge. Adversarial data occupy a distinctly different region compared to natural data. The adversarial sample tends to occupy a steep, narrow region, whereas a typical natural data sample is characterized by a flatter, wider region \cite{zheng2023detecting}. Such behavior results from a small perturbation introduced by the attack that pushes the sample just across the decision boundary of the model. This perturbation results in local minima in the loss landscape, which manifest as steep, narrow regions. We can exploit this behavior to create an effective method for distinguishing between adversarial and natural samples. However, establishing a metric to interpret low and high sharpness remains challenging.

The main idea of this research is to illustrate that the Intrinsic Dimension (ID) can serve as a criterion to interpret low and high sharpness. Consequently, we introduce a general method to detect adversarial samples using the ID as a criterion. By comparing our method's accuracy with other adversarial detection techniques, we achieve a significant improvement in accuracy. These results indicate that the perturbation introduced to natural data also affects the sharpness of the ID, and this impact can be used to accurately detect adversarial data.

Our main contributions are: (1) introducing ID as a quantitative metric to indicate landscape sharpness, thereby enabling the exploitation of such a metric in adversarial detection; (2) designing an algorithm based on ID to detect adversarial input samples in both batch and single formats; and (3) demonstrating that our adversarial sample detection model outperforms traditional adversarial detection models against various types of adversarial attacks.

% ----------------------------------------------------------------

\section{Related Works}

Intrinsic dimensionality (ID) expresses the minimal degrees of freedom required to describe data locally, representing the dimension of the manifold on which the data concentrate. High-ID regions are typically diffuse and fragile, while low-ID regions are compact and stable.

The first robustness-centred treatment of ID was given by Amsaleg et al. \cite{amsaleg2017vulnerability}. Using an MLE over nearest-neighbour distances, they proved that for $k$-NN classifiers, the expected perturbation needed to flip a label is inversely proportional to ID. This result establishes a formal link between geometric complexity and adversarial vulnerability: as ID grows, a sample's protective margin shrinks.

Ansuini et al. \cite{ansuini2019intrinsic} extended this idea to deep networks. By applying TwoNN and MLE estimators to activations from major CNN architectures, they found that ID typically peaks in mid-network layers and declines towards the classifier head. Crucially, they showed that a lower final-layer ID correlates with higher test accuracy, positioning ID as a diagnostic for representation quality and confirming it can be measured reliably in modern networks.

Building on these foundations, Ma et al. \cite{ma2018characterizing} introduced Local Intrinsic Dimensionality (LID), a point-wise estimator based on extreme-value theory. They reported a consistent gap where adversarial examples exhibit higher LID than natural inputs in hidden layers. This property was leveraged to construct an LID-based detector with strong performance against white-box attacks. The method, however, requires access to intermediate activations, limiting its practicality at inference time.

Taken together, these studies provide a theoretical justification for ID's role in robustness and demonstrate that feature-space ID can separate natural from adversarial data. However, they operate on inputs or activations, leaving open whether the parameter-gradient space carries an equally discriminative signal. This question motivates our work.

Gradient- and Loss-Geometry Approaches. A complementary line of research focuses on signals from the input-loss landscape and its gradients. Huang et al. \cite{huang2021gradnorm} proposed GradNorm, which uses the $l_2$-norm of parameter gradients as a detector, where larger norms indicate adversarial inputs. While simple, GradNorm produces a scalar score and can be sensitive to noise, motivating richer geometric descriptors.

\begin{figure}[t]
  \centering
  \includegraphics[width=0.75\linewidth]{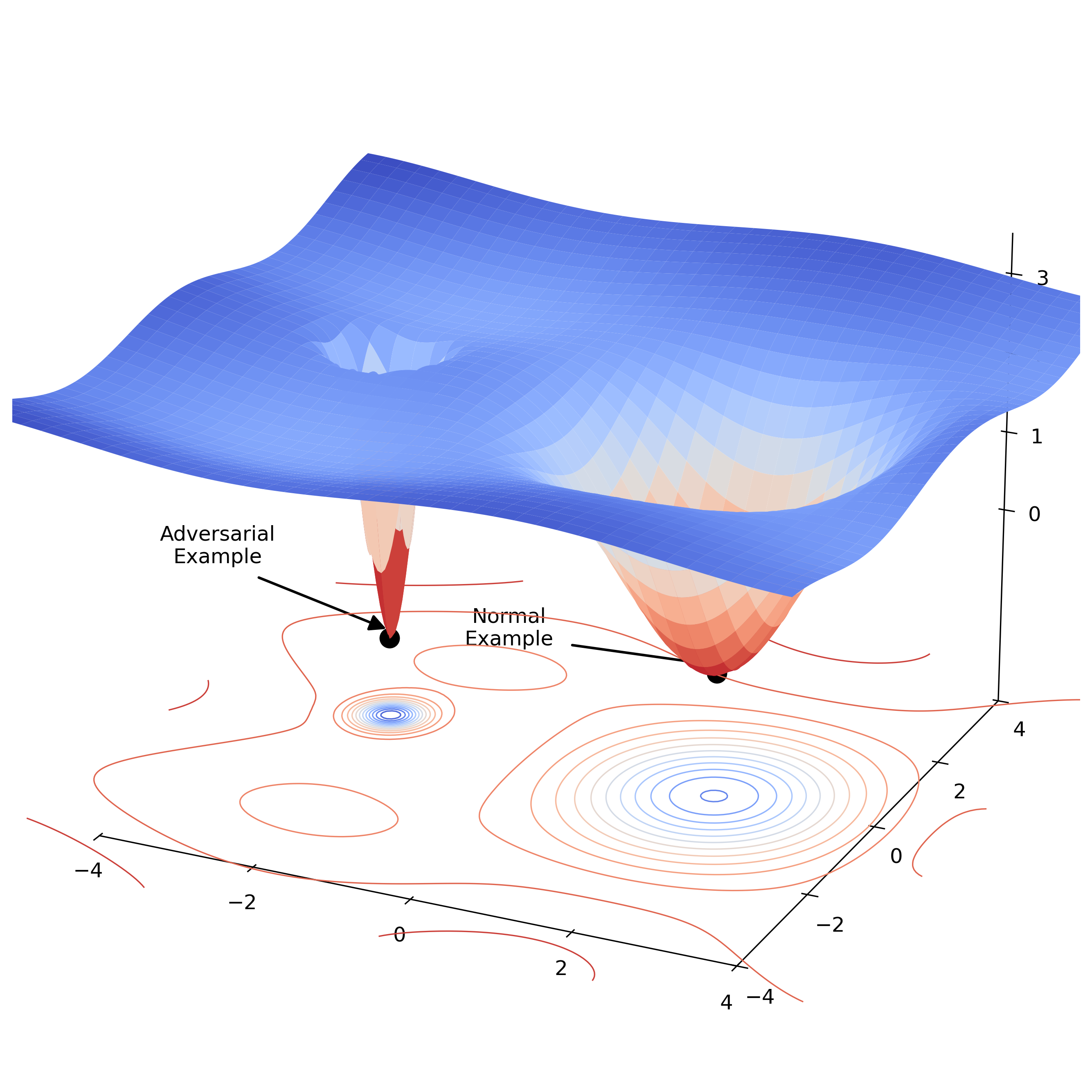}
  \caption{A visualization of the input loss landscape illustrating the geometric difference between natural and adversarial examples. Adversarial examples typically lie in sharp, narrow minima, whereas natural examples are found in wider, flatter basins. This visualization is adapted from the concept presented by Zheng et al. \cite{zheng2023detecting}.}
  \label{fig:loss_landscape}
\end{figure}

Zheng et al. \cite{zheng2023detecting} adopted a landscape-centric view, showing that adversarial examples inhabit sharp, narrow minima while natural inputs occupy broader, flatter basins, a contrast visualized in Figure~\ref{fig:loss_landscape}. They crafted a sharpness-based detector that estimates local curvature, confirming that landscape geometry encodes a separable signal.

These studies confirm that parameter gradients carry discriminative information and that the loss landscape curvature differs for natural and adversarial data. Our work unifies these perspectives by investigating whether the intrinsic dimensionality of gradient embeddings can capture both the large-norm signals and the sharp-valley geometry, providing a single, powerful criterion for detection.

% ----------------------------------------------------------------

\section{Preliminaries}
While modern datasets often reside in a high-dimensional "ambient" space (e.g., an image as a vector in $\mathbb{R}^n$), their inherent structure is typically confined to a much lower-dimensional manifold. The intrinsic dimension (ID) quantifies the dimensionality of this underlying manifold, representing the minimal number of parameters needed to describe the data's structure without significant information loss. A classic illustration is a crumpled sheet of paper in a room: although any point on the paper requires three coordinates in the 3D ambient space, the points themselves are constrained to a 2D surface, which is its intrinsic dimension. 

Formally, if data points lie on a smooth $d$-dimensional manifold $\mathcal{M}\subset\mathbb{R}^n$, its intrinsic dimension is defined as:
\begin{equation}
\mathrm{ID}(\mathcal{M}) = \dim(\mathcal{M}) = d
\end{equation}
This can also be expressed locally. The local ID at a point $x_i$ is defined based on the scaling of the number of neighboring data points $N_i(r)$ within a small radius $r$ \cite{grassberger1983measuring}:
\begin{equation}
\mathrm{ID}(x_i) = \lim_{r\to 0} \frac{\log N_i(r)}{\log(r)}
\label{eq:local_id_def}
\end{equation}
Since the true ID is unknown for real-world data, it must be estimated from samples. A variety of techniques exist, and in this work, we primarily employ two well-established local estimators detailed in our Methodology: the Maximum Likelihood Estimator (MLE) \cite{levina2004maximum} and the Two-Nearest-Neighbors (TwoNN) algorithm \cite{facco2017estimating}.

The concept of ID provides a powerful lens for analyzing the geometry of adversarial examples. A common hypothesis in prior work is that adversarial perturbations, designed to exploit model sensitivities, effectively create a more complex and "brittle" local geometry around the data point. This is often observed as an increase in the local ID of the input space for adversarial examples compared to their natural counterparts \cite{ma2018characterizing}.

In this work, however, we explore a fundamentally different and consequential effect. Instead of analyzing the input data manifold, we investigate the geometric properties of the parameter-gradient space. Our central thesis posits that the localized sharpness of the loss landscape associated with adversarial examples powerfully constrains the model's response. This constraint forces the corresponding parameter gradients, $\nabla_\theta L(\theta; x, y)$, into a highly correlated and therefore lower-dimensional subspace. This creates a discernible and opposite disparity to what is observed in the input space. We hypothesize that for a set of natural gradient embeddings $G_{\mathrm{natural}}$ and a set of adversarial embeddings $G_{\mathrm{adversarial}}$, their respective intrinsic dimensions will consistently satisfy:
\begin{equation}
  \mathrm{ID}(G_{\mathrm{natural}}) > \mathrm{ID}(G_{\mathrm{adversarial}})
  \label{eq:main_hypothesis}
\end{equation}
This shift in perspective—from the geometry of inputs to the geometry of gradients—forms the foundation of our detection method, providing a clear and quantifiable criterion that requires no modification to the model architecture.

% ----------------------------------------------------------------
\section{Methodology}
\subsection{Conceptual Framework}
The primary objective of our proposed methodology is to differentiate
adversarial inputs from their natural counterparts through an examination
of the geometric influence of minute input perturbations on the model’s
parameter gradients. While traditional detection strategies commonly
prioritize input-space perturbations or confidence scores, we posit that
the intrinsic dimension (ID) of the gradient embedding—defined
as the minimum number of directions essential to capture the predominant
variability in
\(
\nabla_\theta L(\theta; x, y)
\)
—constitutes a robust and model-agnostic signature of adversarial
behavior. Fundamentally, rather than inquiring,
What is the magnitude of the loss alteration induced by
perturbing \(x\)?, our inquiry shifts to,
To what extent are the directions in which the model’s
parameters respond to \(x\) structured or unconstrained?

Drawing upon the sharpness analysis conducted by Zheng et al. \cite{zheng2023detecting}, which demonstrated that adversarial samples typically occupy
narrow, high-curvature minima within the input–loss landscape, we
hypothesize that this localized “spikiness” compels the associated
parameter gradients into a more highly constrained subspace. Intuitively,
if a negligible change in \(x\) can induce a drastic shift in the loss
function, the model is compelled to coordinate its parameter updates
along highly specific axes to maintain performance. Conversely, for
natural samples residing in expansive, flatter regions, the loss
exhibits comparative insensitivity to minor input fluctuations, thereby
permitting the gradient vectors to disperse across a greater multiplicity
of directions.

\begin{equation*}
  \begin{aligned}
    G_{\mathrm{natural}}
    &= \{\nabla_\theta L(\theta; x_i, y_i)\}_{i=1}^N\\
    G_{\mathrm{adv}}
    &= \{\nabla_\theta L(\theta; x_i + \delta_i, y_i)\}_{i=1}^N
  \end{aligned}
  \quad
  \begin{aligned}
    ID_{\mathrm{natural}}
    &= \mathrm{ID}(G_{\mathrm{natural}})\\
    ID_{\mathrm{adv}}
    &= \mathrm{ID}(G_{\mathrm{adv}})
  \end{aligned}
\end{equation*}
Our central hypothesis is therefore
\[
  ID_{\mathrm{natural}} \;>\; ID_{\mathrm{adv}}
\]
thereby furnishing a clear, quantifiable disparity that can be
effectively harnessed for the detection of adversarial manipulations
without necessitating any architectural modifications to the underlying
model or additional training.

\subsection{Gradient Embedding Computation}

Our framework is model-agnostic. For empirical validation, we employ standard architectures such as ResNet-50 and ResNet-18~\cite{He_2016_CVPR}, pretrained on ImageNet. We replaced its final fully connected layer to match our target task and fine-tuned all layers on the clean training set until convergence, employing early stopping on a held-out validation split. This configuration balances strong representational capacity with community familiarity, promoting the generality of our findings across modern vision models.
Let \(f_\theta(x)\) be the network’s softmax output for input \(x\). We define the cross-entropy loss for the true label \(y\) as
\[
  L(\theta; x, y)
  = -\log\bigl[f_\theta(x)_y\bigr]
\]
For each sample \((x,y)\), we compute the full-parameter gradient:
\[
  g(x,y)
  = \nabla_\theta L(\theta; x, y)
\]
yielding a vector in \(\mathbb{R}^P\), where \(P\) is the number of trainable parameters.

In our experiments, to improve computational efficiency, we restrict this gradient computation to only the final fully connected layer (or the last few layers), thereby significantly reducing both memory footprint and runtime without materially affecting the separation between natural and adversarial intrinsic dimensions.  
Our natural dataset \(\{(x_i,y_i)\}_{i=1}^N\) consists of validation images held out from fine‑tuning. Adversarial counterparts are generated according to the attack scenarios described in the Experiments section. For each setup, every natural image \(x_i\) yields a perturbed example \(x_i + \delta_i\) that preserves the true label \(y_i\). We then collect the corresponding gradient embeddings into two sets:
\[
  G_{\mathrm{natural}} = \{\,g(x_i, y_i)\}_{i=1}^N,
  \quad
  G_{\mathrm{adv}}     = \{\,g(x_i + \delta_i, y_i)\}_{i=1}^N
\]
We evaluate our detector in two distinct usage scenarios. In the batch mode, inputs arrive grouped—e.g.\ all samples sent by a single agent over a time window—and we compute and aggregate intrinsic‑dimension statistics across the entire batch to decide whether that agent is launching an adversarial attack. This collective analysis can amplify subtle anomalies but blends individual sample signatures. In the single‑sample mode, each example is processed independently, with its own gradient embedding and ID computed to determine adversariality on a per‑input basis. While batch mode benefits from aggregation and reduced per‑sample overhead, single‑sample mode preserves maximal sensitivity to the local curvature differences that distinguish adversarial perturbations.  

\subsection{Intrinsic Dimension Estimation}

Depending on the experiment, we estimate the intrinsic dimension of either \(G_{\mathrm{natural}}\) or \(G_{\mathrm{adv}}\) using either the Two‑Nearest Neighbors (TwoNN) estimator~\cite{facco2017estimating}, which computes dimension from the ratio of distances to the first and second nearest neighbors, or the Maximum Likelihood Estimator (MLE)~\cite{levina2004maximum}, which applies a log‑ratio formula over \(k\) nearest neighbors.
Following standard practice, TwoNN uses \(m=2\) neighbors, and we subsample a fixed number of gradient embeddings to balance stability and efficiency. For MLE, we choose \(k=10\) and average the estimate over multiple bootstrap samples to reduce variance.
First, we compute all gradient embeddings \(g(x_i,y_i)\) for samples in both sets—restricted to the final (or last few) layers as previously described. Next, for each embedding, we determine its distances to the \(m\) nearest neighbors (\(m=2\) for TwoNN, \(m=k\) for MLE) and apply the corresponding estimation formula: the distance‑ratio moment estimator for TwoNN, and the log‑ratio bootstrap average for MLE. Finally, we aggregate the results to obtain  
\[
  ID_{\mathrm{natural}} = \mathrm{ID}\bigl(G_{\mathrm{natural}}\bigr), 
  \quad
  ID_{\mathrm{adv}}     = \mathrm{ID}\bigl(G_{\mathrm{adv}}\bigr).
\]
Although our gradient embeddings are already confined to the final (or last few) layers, additional efficiency gains are possible. For example, one might estimate intrinsic dimension on a random subset of gradient vectors instead of the entire set; employ approximate nearest‑neighbor search libraries such as FAISS~\cite{johnson2019billion} to accelerate distance computations; or distribute neighbor searches and distance calculations across multiple GPU cores. Each of these approaches can substantially reduce runtime with minimal impact on estimation accuracy. 

% ----------------------------------------------------------------
\begin{figure}[t]
  \centering
  \includegraphics[width=0.9\textwidth]{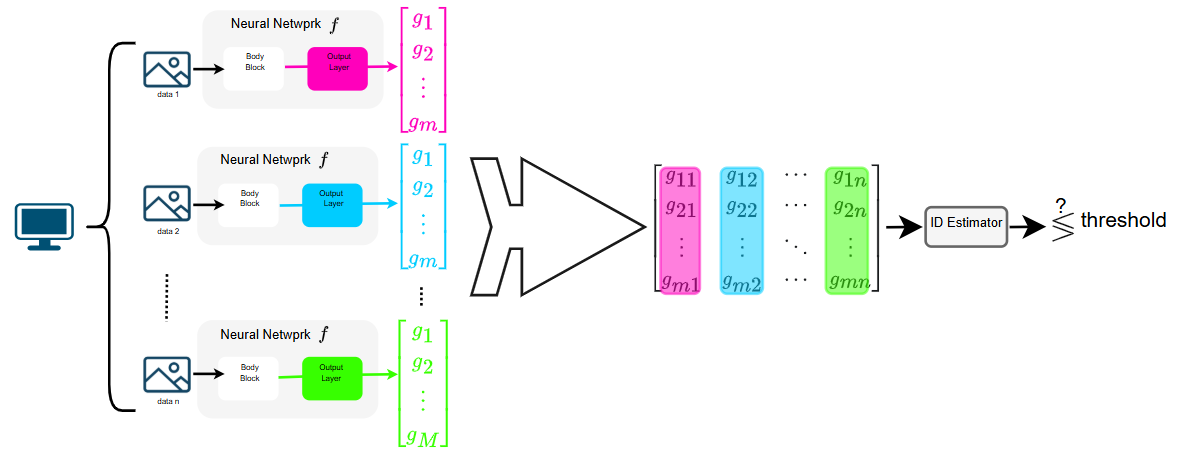}
  \caption{An overview of our batch-wise adversarial detection pipeline. A batch of $n$ input samples is processed by a neural network $f$. For each sample, the gradient vector $g$ of the loss with respect to the output layer's parameters is computed. These $n$ gradient vectors are aggregated into a single set, whose intrinsic dimension (ID) is then estimated. Finally, the resulting ID value is compared against a pre-determined threshold to classify the entire batch as either natural or adversarial.}
  \label{fig:batch}
\end{figure}

\section{Experiment}
To validate our hypothesis, we conduct a series of experiments designed to evaluate our ID-based detection method across diverse datasets, attack methodologies, and operational scenarios.

\subsection{Batch‑Wise Gradient Analysis}

\paragraph{Problem Setting.}  
We consider a setting inspired by Federated Learning \cite{mcmahan2017communication}, where a central server receives gradient updates from \(K\) clients in mini‑batches. Each client \(k\) holds a local dataset  
\[
  D_k=\{(x_{k,i},y_{k,i})\}_{i=1}^{N_k}
\]
and computes its batch of gradient embeddings  
\[
  G_k = \bigl\{\nabla_\theta L(\theta; x_{k,i},y_{k,i})\bigr\}_{i=1}^{N_k}.
\]
Malicious clients may inject adversarial examples \(x_{k,i}^{\mathrm{adv}}\) (e.g.\ via FGSM, PGD, BIM, or DeepFool), yielding  
\[
  G_k^{\mathrm{adv}} = \bigl\{\nabla_\theta L(\theta; x_{k,i}^{\mathrm{adv}},y_{k,i})\bigr\}_{i=1}^{N_k}.
\]
The server maintains a trusted reference set  
\[
  G_{\mathrm{natural}} = \{\nabla_\theta L(\theta; x_i,y_i)\}_{i=1}^{N},
  \quad
  ID_{\mathrm{natural}} = \mathrm{ID}(G_{\mathrm{natural}}).
\]
For each client \(k\), the server computes  
\[
  ID(G_k),\quad
  \Delta_k = \bigl|\,ID(G_k) - ID_{\mathrm{natural}}\bigr|
\]
and classifies \(k\) as adversarial if \(\Delta_k > \tau\).

\begin{algorithm}[h]
\caption{Adversarial Client Detection via Intrinsic Dimensionality}
\label{alg:fl_adv_detection}
\textbf{Input}: Global model \(f_\theta\), loss \(L(\theta;x,y)\), client datasets \(\{D_k\}_{k=1}^K\), estimator \(\mathrm{estimate\_id}\), threshold \(\tau\), reference dataset \(D_{\mathrm{natural}}\)\\
\textbf{Output}: Client labels \(\{\mathrm{ClientType}_k\}\)
\begin{algorithmic}[1]
  \STATE \(G_{\mathrm{natural}}\gets \{\nabla_\theta L(\theta;x_i,y_i)\}_{i=1}^N\)
  \STATE \(ID_{\mathrm{natural}}\gets \mathrm{estimate\_id}(G_{\mathrm{natural}})\)
  \FOR{\(k = 1\) to \(K\)}
    \STATE \(G_k\gets \{\nabla_\theta L(\theta;x_{k,i},y_{k,i})\}_{i=1}^{N_k}\)
    \STATE \(ID_k\gets \mathrm{estimate\_id}(G_k)\)
    \STATE \(\Delta_k\gets |ID_k - ID_{\mathrm{natural}}|\)
    \IF{\(\Delta_k \le \tau\)}
      \STATE \(\mathrm{ClientType}_k\gets\) natural
    \ELSE
      \STATE \(\mathrm{ClientType}_k\gets\) adversarial
    \ENDIF
  \ENDFOR
  \STATE \textbf{return} \(\{\mathrm{ClientType}_k\}\)
\end{algorithmic}
\end{algorithm}

\paragraph{Dataset and Experimental Setup.}  
We simulate \(K=5\) clients on three benchmarks—SVHN \cite{netzer2011reading}, MNIST \cite{lecun1998gradient}, and CIFAR‑10 \cite{krizhevsky2009learning}—using a ResNet‑50 model (gradients computed on the final two layers). Client 1 sends a clean mini‑batch of size \(N_1\); each of the remaining four clients applies one of FGSM \cite{goodfellow2014explaining}, PGD \cite{madry2017towards}, BIM \cite{kurakin2016adversarial}, or DeepFool \cite{moosavi2016deepfool} (\(\epsilon=8/255\), PGD/BIM: 10 steps, step \(2/255\)). The server fixes \(\tau\) by cross‑validation on held‑out natural vs.\ adversarial gradient batches.

\begin{figure}[htbp]
    \centering
    \begin{subfigure}[b]{0.48\textwidth}
        \includegraphics[width=\linewidth]{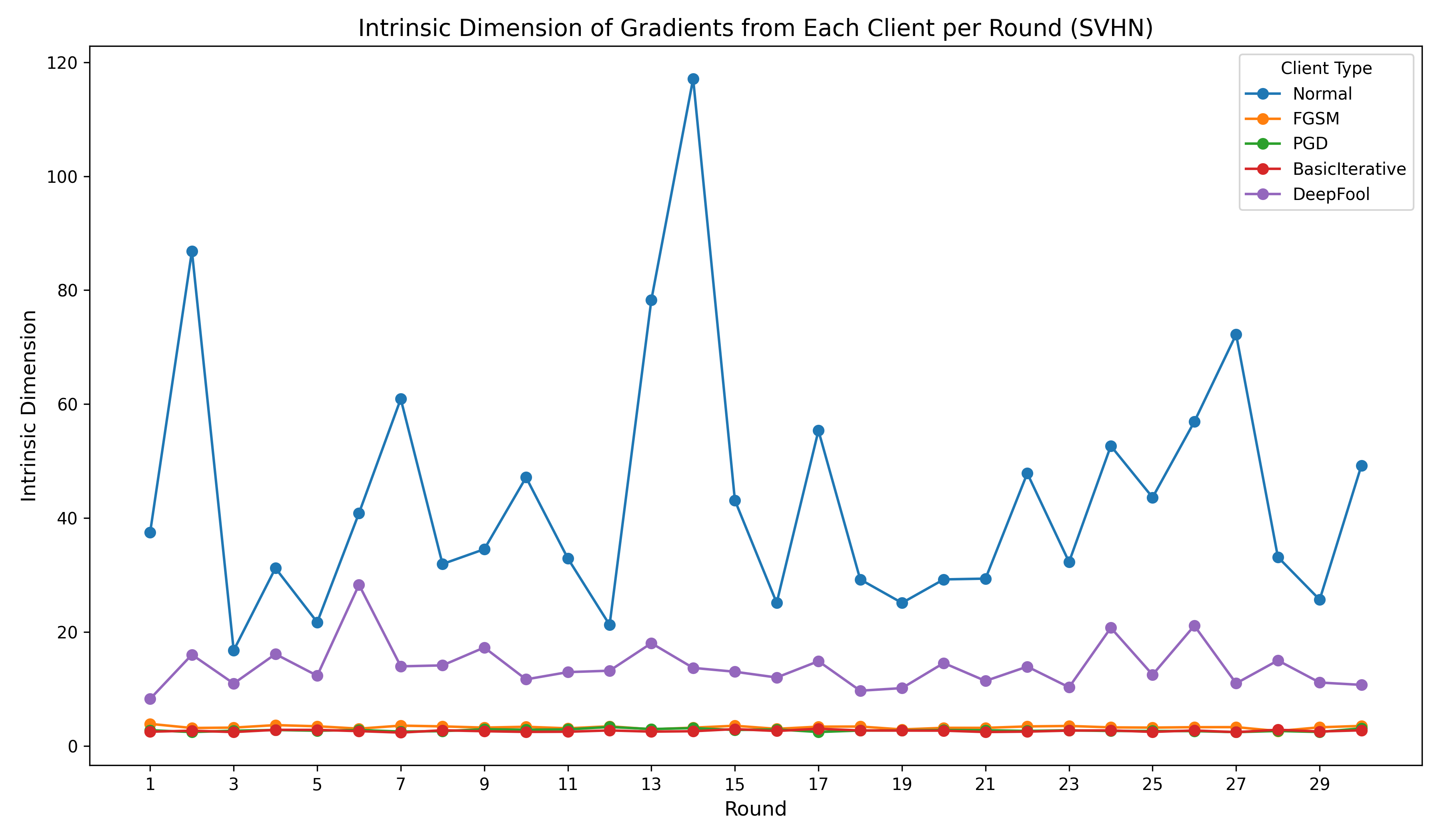}
        \caption{Raw ID Values (SVHN)}
    \end{subfigure}
    \hfill 
    \begin{subfigure}[b]{0.48\textwidth}
        \includegraphics[width=\linewidth]{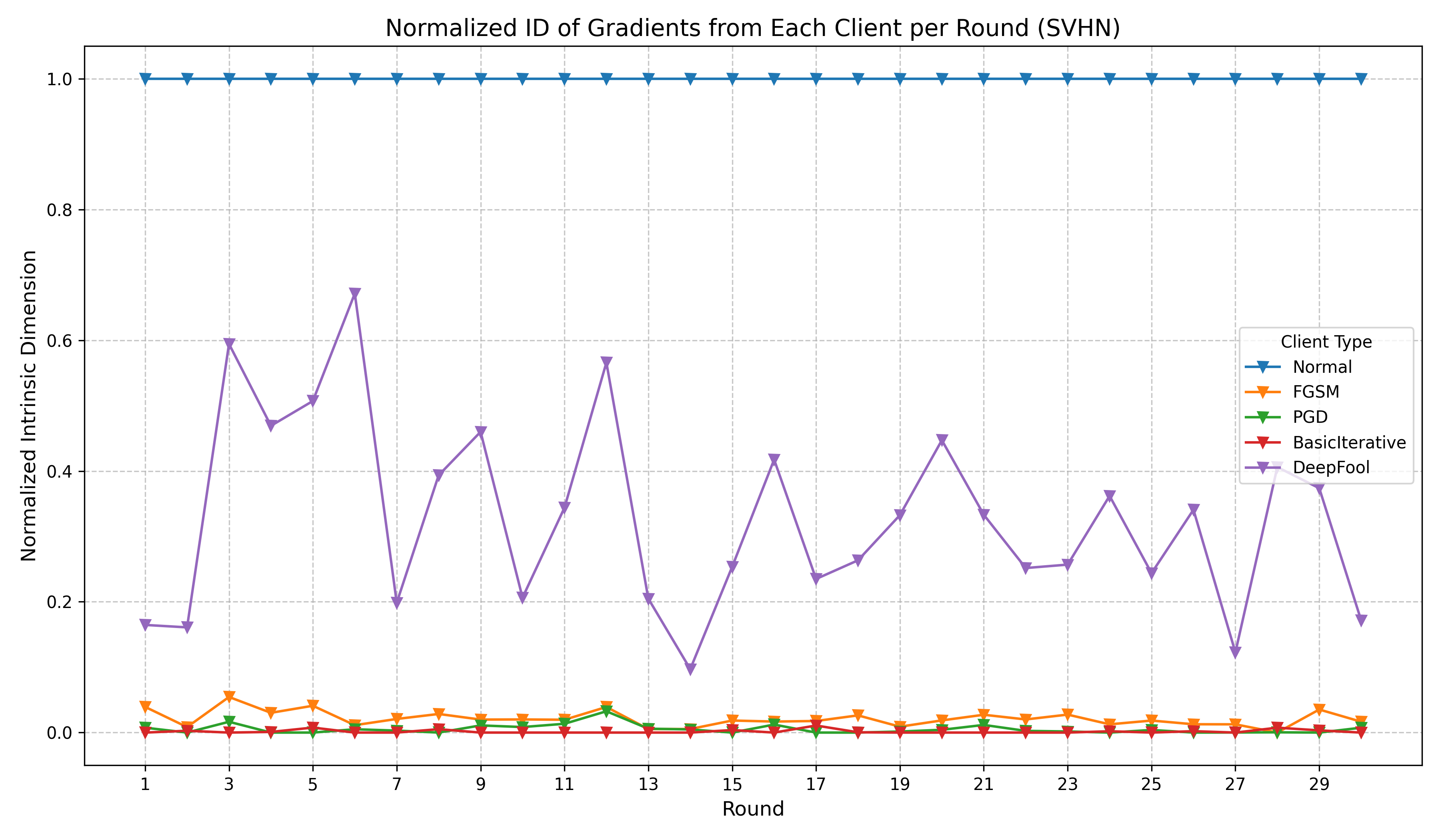}
        \caption{Normalized ID Values (SVHN)}
    \end{subfigure}
    \caption{Batch-wise detection results on SVHN. The natural client (Natural) consistently exhibits a different intrinsic dimension from the four malicious clients using various attacks, enabling robust detection. Full results for other datasets are in the Appendix.}
    \label{fig:batch_id_results_main}
\end{figure}

\paragraph{Compared Methods and Results.}  
We compare our ID Detector (Algorithm~\ref{alg:fl_adv_detection}) against two common baselines: an Average-Gradient Norm detector and a Confidence Filtering method. Figure~\ref{fig:batch_id_results_main} illustrates the results for the SVHN dataset, showing a clear and consistent separation between the intrinsic dimension of the benign client's gradients and those of the four malicious clients. This distinct gap allows for highly accurate detection, achieving over 95\% accuracy in our simulations. In contrast, the baseline methods often failed to distinguish the clients due to significant overlap in their respective metrics. This trend holds across all tested datasets; full results for MNIST and CIFAR-10, which show similar patterns of separability, are provided in the Appendix (Figures~\ref{fig:appendix_mnist_results} and \ref{fig:appendix_cifar10_results}).

\subsection{Individual Gradient Analysis}
\begin{figure}[t]
  \centering
  \includegraphics[width=\textwidth]{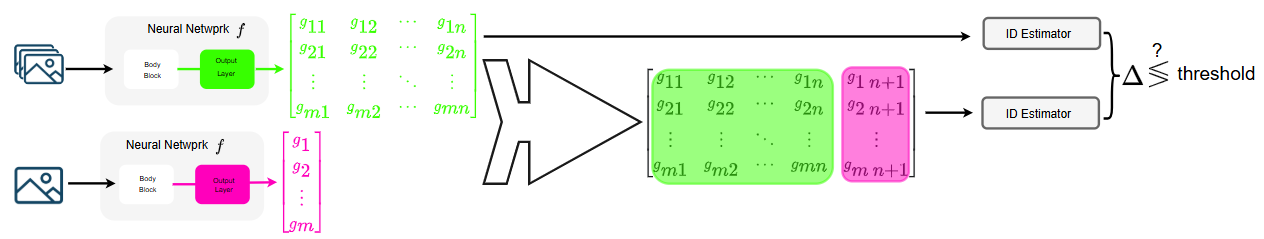}
  \caption{Workflow for per-sample adversarial detection using differential intrinsic dimension analysis. The method establishes a baseline intrinsic dimension ($ID_{\mathrm{natural}}$) from a reference set of $n$ natural gradient embeddings (top, green). For an incoming test sample (bottom, magenta), its gradient embedding is appended to the reference set to form an augmented manifold. The ID of this augmented set ($ID_{\mathrm{aug}}$) is then computed. The sample is classified as adversarial if the change in ID ($\Delta = |ID_{\mathrm{aug}} - ID_{\mathrm{natural}}|$) surpasses a defined threshold, indicating a significant geometric disruption.}
  \label{fig:individual}
\end{figure}

\paragraph{Problem Setting.}  
For safety-critical applications such as autonomous driving, real-time detection of individual adversarial samples is paramount. Our workflow for this scenario is depicted in Figure~\ref{fig:individual}. The core idea is to maintain a reference manifold of natural gradient embeddings, $G_{\mathrm{norm}}$, and classify an incoming sample $x^*$ based on how its gradient $g^*$ perturbs the geometry of this manifold. As detailed in Algorithm~\ref{alg:id_eval_percentile}, we classify a sample as adversarial if the ID of the augmented set, $\mathrm{ID}(G_{\mathrm{norm}}\cup\{g^*\})$, deviates significantly from the baseline $\mathrm{ID}(G_{\mathrm{norm}})$, which we measure using percentile-based thresholds.
 
\begin{algorithm}[h]
\caption{Per‑Sample Adversarial Detection via ID (Percentile)}
\label{alg:id_eval_percentile}
\textbf{Input}: Model \(f_\theta\), loss \(L\), reference batch \(B_{\mathrm{norm}}\), test sample \((x^*,y^*)\), estimator \(\mathrm{estimate\_id}\)\\
\textbf{Parameter}: Percentiles \(P_{10},P_{90}\) from \(G_{\mathrm{norm}}\)\\
\textbf{Output}: \(\mathrm{Label}\in\{\mathrm{natural},\mathrm{adversarial}\}\)
\begin{algorithmic}[1]
  \STATE Compute \(G_{\mathrm{norm}}\gets\{\nabla_\theta L(\theta;x_i,y_i)\}\) for \((x_i,y_i)\in B_{\mathrm{norm}}\)
  \STATE Compute \(ID_{\mathrm{norm}}\gets\mathrm{estimate\_id}(G_{\mathrm{norm}})\)
  \STATE Compute \(g^*\gets\nabla_\theta L(\theta;x^*,y^*)\)
  \STATE \(G_{\mathrm{aug}}\gets G_{\mathrm{norm}}\cup\{g^*\}\)
  \STATE \(ID_{\mathrm{aug}}\gets\mathrm{estimate\_id}(G_{\mathrm{aug}})\)
  \IF{\(ID_{\mathrm{aug}}\in[P_{10},P_{90}]\)}
    \STATE \textbf{return} natural
  \ELSE
    \STATE \textbf{return} adversarial
  \ENDIF
\end{algorithmic}
\end{algorithm}

\paragraph{Dataset and Experimental Setup.}  
We evaluate on SVHN \cite{netzer2011reading} using a pretrained ResNet‑18, extracting gradients from its final fully connected layer. We form \(B_{\mathrm{norm}}\) by randomly selecting \(N=50\) natural test samples. Adversarial examples are generated via PGD \cite{madry2017towards} (\(\epsilon=8/255\), 10 steps, step size \(2/255\)) and AutoAttack \cite{croce2020reliable} (default settings). Percentiles \(P_{10},P_{90}\) are computed over \(ID(G_{\mathrm{norm}})\).

\begin{figure}[htbp]
    \centering
    \begin{subfigure}[b]{0.48\textwidth}
        \includegraphics[width=\linewidth]{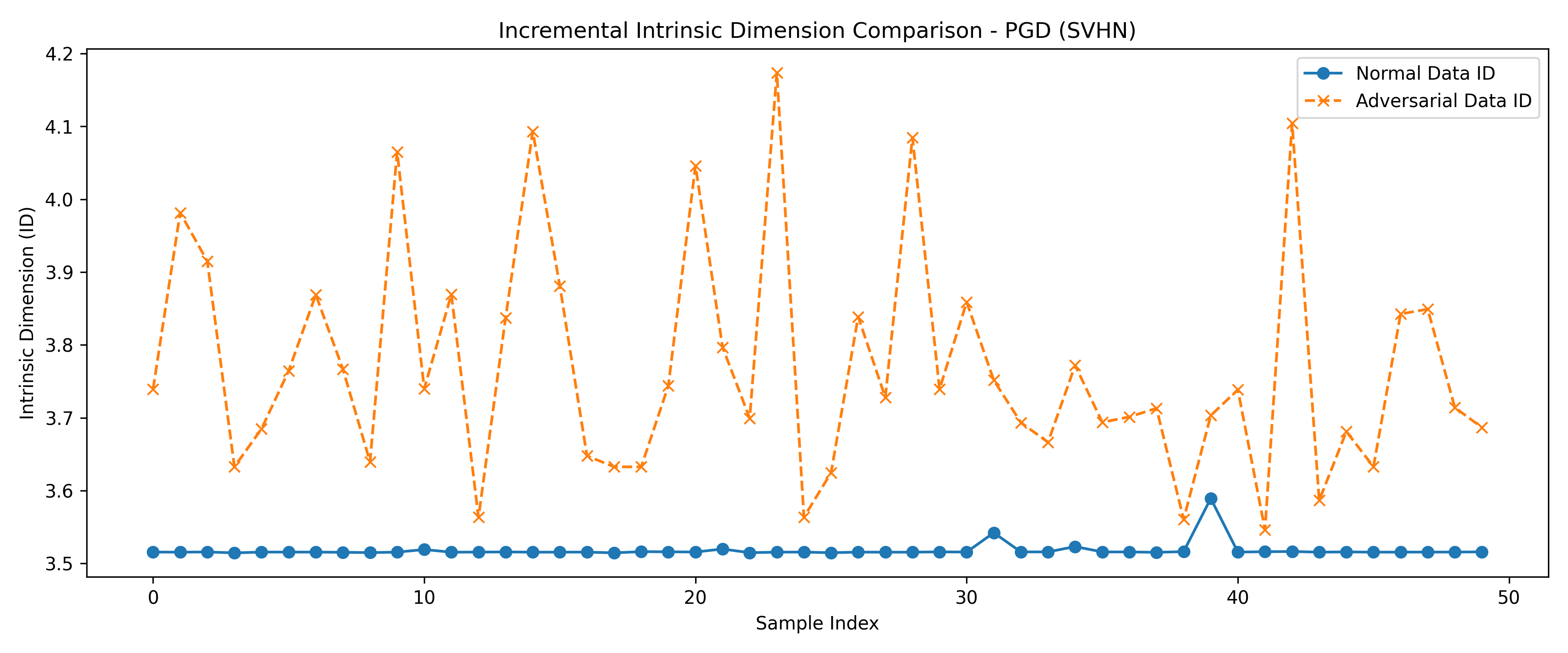}
        \caption{PGD}
    \end{subfigure}
    \hfill
    \begin{subfigure}[b]{0.48\textwidth}
        \includegraphics[width=\linewidth]{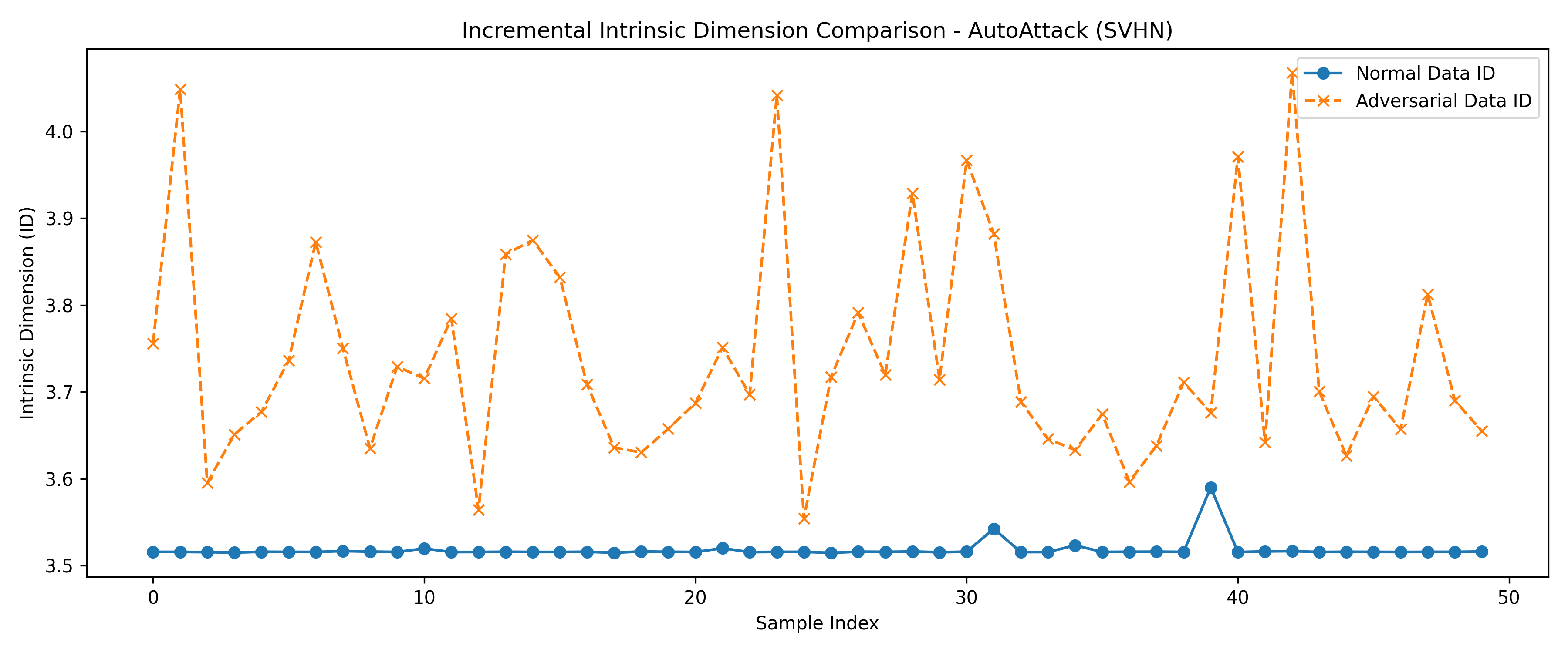}
        \caption{AutoAttack}
    \end{subfigure}
    \caption{Intrinsic dimensionality analysis on SVHN. Intrinsic dimensionality of the gradient embeddings deviates significantly when adversarial perturbations are applied, supporting the effectiveness of the proposed detection method.}
    \label{fig:individual_id_results}
\end{figure}

\begin{figure}[h]
    \centering
    \includegraphics[width=0.7\linewidth]{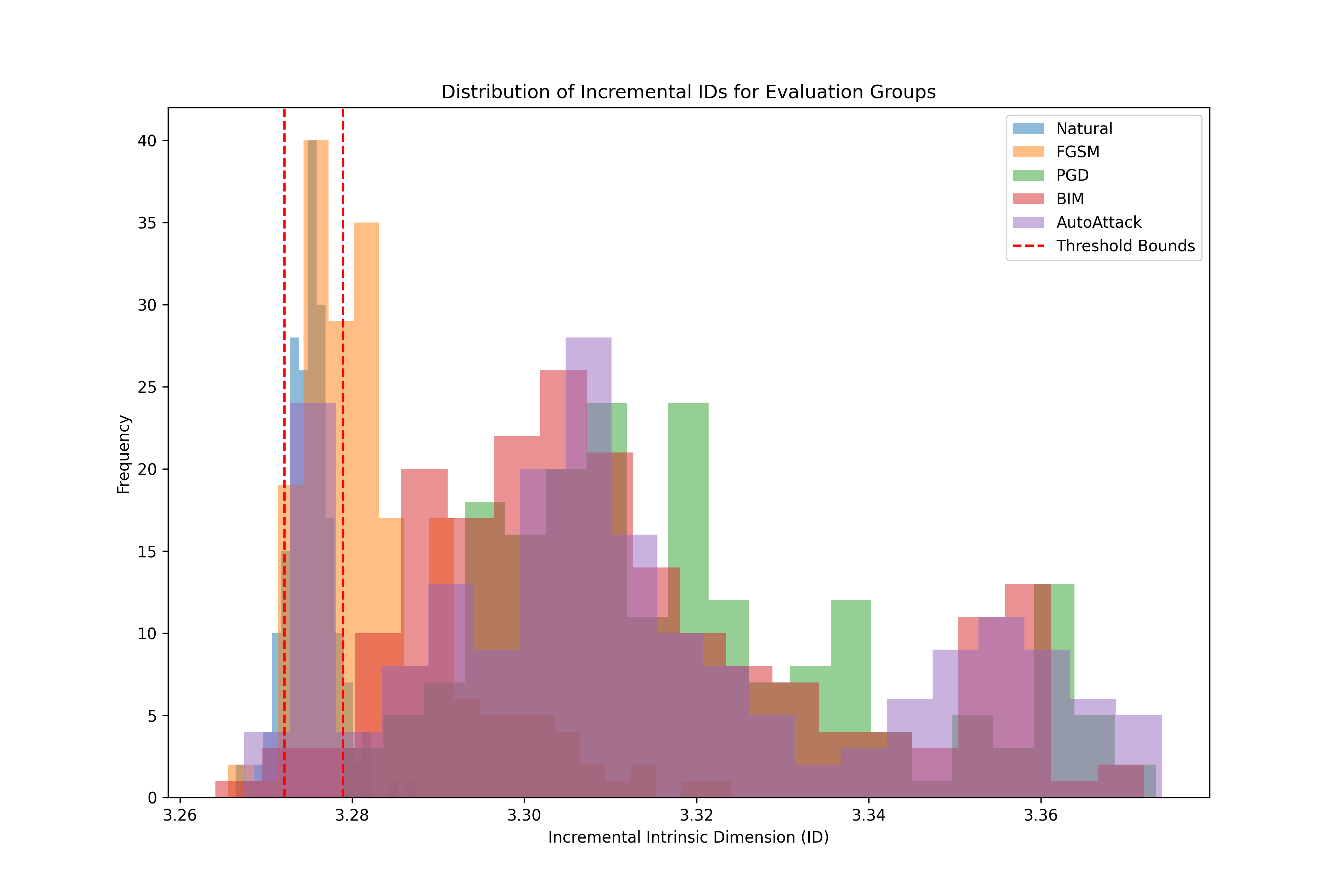}
    \caption{Distribution of incremental ID values on SVHN. Adversarial attacks consistently induce a distributional shift relative to natural data, allowing the percentile-based thresholds (dashed lines) to effectively separate them.}
    \label{fig:svhn_histogram}
\end{figure}

\paragraph{Compared Methods and Results.}
We compare our ID-percentile detector against two baselines: a Gradient-Norm Threshold and a Confidence-Score Filter. On SVHN, our method achieves a strong 85.4\% overall detection accuracy. This effectiveness stems from the clear geometric separability between the incremental ID of natural and adversarial samples, a pattern visualized for PGD and AutoAttack in Figure~\ref{fig:individual_id_results}. Furthermore, the ID distributions in Figure~\ref{fig:svhn_histogram} confirm that our percentile-based thresholds effectively demarcate benign gradients from those generated by a range of attacks.

\subsection{Comparison on CIFAR‑10 \& COCO}
\paragraph{Datasets and Attacks.}
We evaluate our per-sample intrinsic-dimension detector on two standard benchmarks: the 10,000 test images of \textbf{CIFAR-10} \cite{krizhevsky2009learning} and a curated subset of 4,952 single-label images from \textbf{MS COCO 2017} \cite{lin2014microsoft}. For both datasets, we hold out 1,000 clean images for threshold calibration and use the remainder for the final evaluation. Adversarial examples are generated using the five widely adopted attacks detailed in Table~\ref{tab:attack_params}.

\begin{table}[tb]
  \centering
  \small
  \begin{tabular}{l|l}
    \toprule
    \textbf{Attack}       & \textbf{Parameters}                                    \\
    \midrule
    FGSM                  & $\epsilon=0.008$                                       \\
    PGD                   & $\epsilon=0.01$, $\alpha=0.02$, steps=40               \\
    BIM                   & $\epsilon=0.03$, $\alpha=0.01$, steps=10               \\
    DeepFool              & steps=20                                               \\
    CW (L$_2$)            & $C=2$, $\kappa=2$, steps=500, lr=0.01                  \\
    \bottomrule
  \end{tabular}
  \caption{Adversarial Attack Parameters}
  \label{tab:attack_params}
\end{table}

\paragraph{Experimental Setup.}
For CIFAR-10, we fine-tune a ResNet-18, while for MS COCO, we train only a linear head atop a frozen, pretrained ResNet-18 backbone. In all experiments, we extract gradients from the final fully-connected layer. For CIFAR-10, we apply the MLE estimator~\cite{levina2004maximum} to the gradient embeddings after reducing them to 10 dimensions via PCA. For MS COCO, we directly apply the TwoNN estimator~\cite{facco2017estimating} to the original embeddings. Detection thresholds are calibrated across both datasets by selecting the 10th and 90th percentiles of incremental ID values derived from the 1,000 held-out clean samples. 
We report the adversarial detection rate  
\[
  \mathrm{DR}_{\mathrm{a}} = \frac{\mathrm{TP}}{\mathrm{TP} + \mathrm{FN}}\times100\%
\]  
where TP and FN are true positives and false negatives on the adversarial test set.

\paragraph{Results and Discussion}  
Tables~\ref{tab:cifar10_comparison} and \ref{tab:coco_comparison} compare our detection method against nine state‑of‑the‑art methods reproduced from \cite{baselinePaper}. On CIFAR‑10, our method achieves near‑perfect detection on PGD and CW, and exceeds 92\% on all attacks. On COCO, detection rates range from 85.4\% (DeepFool) to 95.3\% (PGD), demonstrating robustness even on high‑resolution data.

\begin{table}[htbp]
    \centering
    \small
    \begin{minipage}{0.48\linewidth}
        \centering
        \caption{Adversarial Detection Rate (\%) on CIFAR‑10}
        \label{tab:cifar10_comparison}
        \resizebox{\linewidth}{!}{
            \setlength{\tabcolsep}{2pt}
            \begin{tabular}{l|ccccc}
                \toprule
                Method & FGSM & PGD & BIM & DF & CW \\
                \midrule
                SAC\cite{SAC32} & 60.1 & 59.7 & 56.8 & 21.6 & 17.7 \\
                Sim‑DNN\cite{SimDNN48} & 70.5 & 60.0 & 49.4 & 26.7 & 22.9 \\
                DTBA\cite{DTBA33} & 78.3 & 75.6 & 71.7 & 36.2 & 32.3 \\
                MH‑UI\cite{MHUI34} & 79.2 & 76.5 & 74.6 & 49.1 & 52.5 \\
                AAE\cite{AAE35} & 80.5 & 76.9 & 75.4 & 63.7 & 60.2 \\
                HSJ\cite{HSJ37} & 77.5 & 75.2 & 75.6 & 60.1 & 59.4 \\
                HM\cite{HM36} & 86.9 & 84.5 & 84.0 & 80.6 & 77.9 \\
                FCB\cite{FCB38} & 49.8 & 47.1 & 43.6 & 15.1 & 11.4 \\
                MF\cite{MF40} & 51.4 & 48.0 & 46.1 & 16.4 & 11.9 \\
                MADM\cite{MADM39} & 62.4 & 54.2 & 51.5 & 19.0 & 14.3 \\
                \midrule
                \textbf{Ours} & \textbf{96.4} & \textbf{100.0} & \textbf{98.4} & \textbf{92.7} & \textbf{100.0} \\
                \bottomrule
            \end{tabular}
        }
    \end{minipage}
    \hfill 
    \begin{minipage}{0.48\linewidth}
        \centering
        \caption{Adversarial Detection Rate (\%) on MS COCO}
        \label{tab:coco_comparison}
        \resizebox{\linewidth}{!}{
            \setlength{\tabcolsep}{2pt}
            \begin{tabular}{l|ccccc}
                \toprule
                Method & FGSM & PGD & BIM & DF & CW \\
                \midrule
                SAC\cite{SAC32} & 58.7 & 56.3 & 37.8 & 21.1 & 16.8 \\
                Sim‑DNN\cite{SimDNN48} & 63.5 & 74.1 & 37.8 & 24.2 & 22.8 \\
                DTBA\cite{DTBA33} & 74.6 & 79.8 & 37.8 & 34.0 & 31.6 \\
                MH‑UI\cite{MHUI34} & 76.0 & 80.3 & 37.5 & 47.5 & 50.1 \\
                AAE\cite{AAE35} & 77.3 & 82.0 & 69.1 & 56.5 & 55.1 \\
                HSJ\cite{HSJ37} & 76.6 & 73.9 & 68.3 & 58.8 & 57.5 \\
                HM\cite{HM36} & 85.6 & 89.6 & 84.9 & 78.3 & 75.8 \\
                FCB\cite{FCB38} & 46.7 & 51.1 & 14.4 & 14.3 & 10.5 \\
                MF\cite{MF40} & 48.8 & 56.4 & 17.7 & 15.9 & 11.5 \\
                MADM\cite{MADM39} & 61.2 & 64.8 & 22.8 & 18.5 & 14.0 \\
                \midrule
                \textbf{Ours} & \textbf{93.9} & \textbf{95.3} & \textbf{86.2} & \textbf{85.4} & \textbf{87.6} \\
                \bottomrule
            \end{tabular}
        }
    \end{minipage}
\end{table}
Our method consistently outperforms existing detectors by a wide margin on CIFAR‑10, achieving perfect detection on three attacks and $\geq 92.7\%$ on all others. On COCO, performance is slightly lower—reflecting higher image complexity—but remains strong $(\geq 85.4\%)$ across all attacks, demonstrating the generality and robustness of intrinsic‑dimension‑based detection in both low‑ and high‑resolution settings.

% ----------------------------------------------------------------

\section{Conclusion}
We introduced a novel adversarial detection method based on the geometry of the model's parameter-gradient space. Our central hypothesis, confirmed through extensive experiments, is that gradients from adversarial examples occupy a manifold of significantly lower intrinsic dimension (ID) than those from natural examples. Our detection pipeline, operating in both batch-wise and single-sample modes, leverages this disparity to establish a new state-of-the-art on benchmarks including CIFAR-10 and MS COCO, achieving over 92\% detection rate on CIFAR-10 against a wide range of white-box attacks.

Despite its strong performance, our method has limitations, including the computational overhead of the single-sample mode and a primary focus on non-adaptive attacks. Future work will concentrate on two critical areas: first, developing more efficient ID estimation techniques for real-time applications, and second, enhancing robustness against powerful adaptive attacks specifically designed to evade our geometric criterion.

% =========================================================
% References
% =========================================================
\begingroup
\raggedright 
\sloppy      
\bibliographystyle{unsrtnat} 
\bibliography{references}
\endgroup

\newpage 

% =========================================================
% Appendix
% =========================================================
\appendix
\section{Appendix}
\label{sec:appendix}

In this appendix, we provide further theoretical details on the intrinsic dimension estimators used in our work and present supplementary results for our batch-wise detection experiments.

\subsection{Theoretical Foundations of Intrinsic Dimension Estimators}
\label{sec:appendix_estimators}
In this section, we provide a more detailed theoretical background for the two local intrinsic dimension (ID) estimators employed in our study. The choice of estimator is critical, as each offers a different trade-off between statistical accuracy, computational cost, and robustness to high ambient dimensions.

\paragraph{Maximum Likelihood Estimator (MLE).}
The MLE, proposed in the seminal work of Levina and Bickel \cite{levina2004maximum}, is a foundational method for ID estimation rooted in statistical likelihood. Its derivation rests on two key assumptions about the data's local geometry:
\begin{enumerate}
    \item The data points are drawn from a probability distribution that is approximately uniform within a small neighborhood of any given point $x_i$.
    \item The number of data points falling within a hypersphere of radius $r$ centered at $x_i$ follows a homogeneous Poisson process.
\end{enumerate}
Under these assumptions, if the data locally lies on a $d$-dimensional manifold, the probability of finding the nearest neighbor outside a radius $r$ decays exponentially with the volume of the hypersphere, which scales as $r^d$.

Let $T_{i,1} < T_{i,2} < \dots < T_{i,k}$ be the ordered Euclidean distances from a point $x_i$ to its $k$ nearest neighbors. By maximizing the joint likelihood of observing these specific distances, one can derive a simple and elegant closed-form estimator for the local intrinsic dimension $\hat{d}$ at point $x_i$:
\begin{equation}
  \hat{d}_{\mathrm{MLE}}(x_i) = \left( \frac{1}{k-1} \sum_{j=1}^{k-1} \log \frac{T_{i,k}}{T_{i,j}} \right)^{-1}
  \label{eq:mle_appendix}
\end{equation}
A global ID estimate for a set of embeddings is then typically computed by averaging these local estimates. The choice of the hyperparameter $k$ is crucial: small values lead to high variance, while large values can introduce bias if the manifold's curvature is non-negligible within the $k$-neighborhood. Due to its proven statistical accuracy on well-behaved, lower-dimensional manifolds, we selected MLE for our experiments on CIFAR-10, where the PCA pre-processing step projects gradients onto a stable, low-dimensional subspace. For these experiments, we use a standard value of $k=10$.

\paragraph{Two-Nearest-Neighbors (TwoNN) Estimator.}
The TwoNN estimator, introduced by Facco et al. \cite{facco2017estimating}, is a more recent, parameter-free method derived from the principles of extreme value theory. It is specifically designed to be robust and efficient, even when the ambient dimension is very high. The core insight of TwoNN is that for any continuous probability distribution, the ratio of distances from a point $x_i$ to its second nearest neighbor ($T_{i,2}$) and its first nearest neighbor ($T_{i,1}$) follows a Pareto distribution. The shape parameter of this distribution is precisely the intrinsic dimension $d$.

This leads to an extremely simple yet powerful local estimator that relies only on the first two neighbors:
\begin{equation}
  \hat{d}_{\mathrm{TwoNN}}(x_i) = \frac{1}{\log(T_{i,2} / T_{i,1})}
  \label{eq:twonn_appendix}
\end{equation}
As with MLE, a global estimate is obtained by averaging the local $\hat{d}_{\mathrm{TwoNN}}(x_i)$ values. The primary advantages of TwoNN are its computational simplicity (requiring only two neighbors) and its lack of tuning parameters like $k$. This makes it particularly suitable for large-scale, high-dimensional datasets where hyperparameter tuning is infeasible and computational cost is a major concern. These properties motivated our choice of TwoNN for the experiments on the high-resolution MS COCO dataset, where the gradient embeddings reside in a much higher-dimensional ambient space.

\subsection{Supplementary Batch-Wise Detection Results}
\begin{figure}[htbp]
    \centering
    \begin{subfigure}[b]{0.48\textwidth}
        \includegraphics[width=\linewidth]{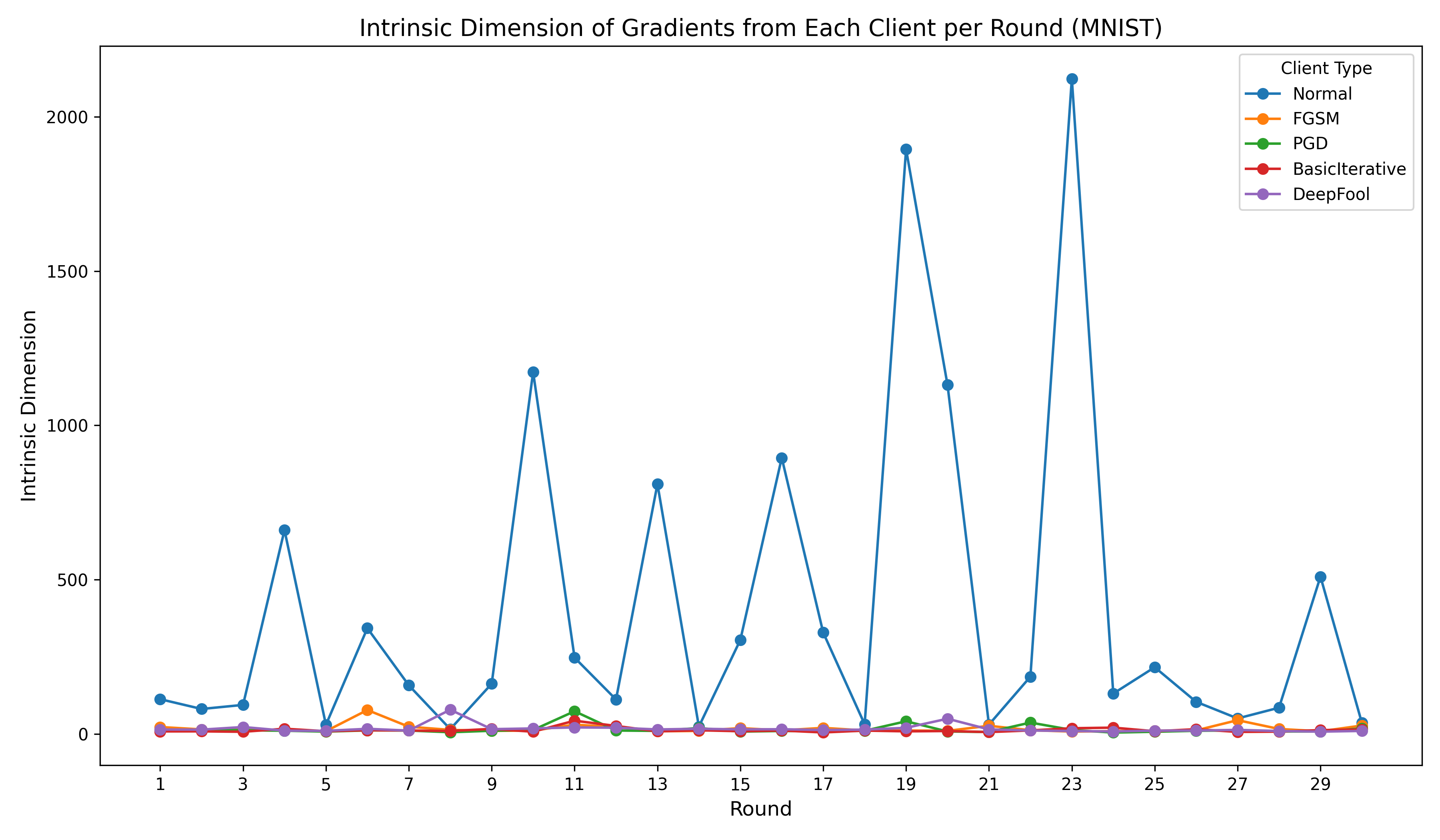}
        \caption{Raw ID Values (MNIST)}
    \end{subfigure}
    \hfill
    \begin{subfigure}[b]{0.48\textwidth}
        \includegraphics[width=\linewidth]{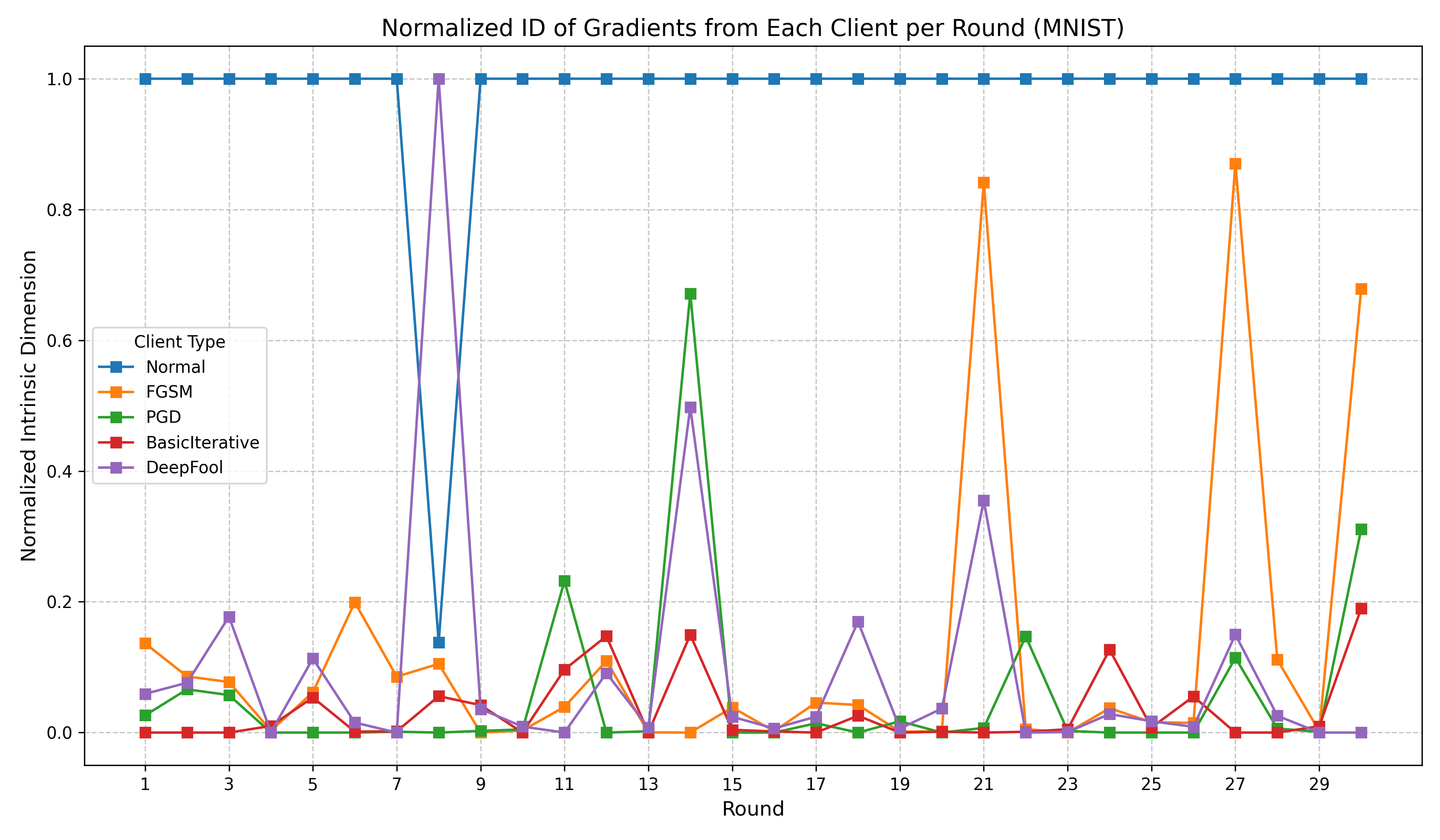}
        \caption{Normalized ID Values (MNIST)}
    \end{subfigure}
    \caption{Batch-wise detection results on MNIST. Clear separability is observed between the benign (Natural) client and malicious clients.}
    \label{fig:appendix_mnist_results}
\end{figure}

\begin{figure}[htbp]
    \centering
    \begin{subfigure}[b]{0.48\textwidth}
        \includegraphics[width=\linewidth]{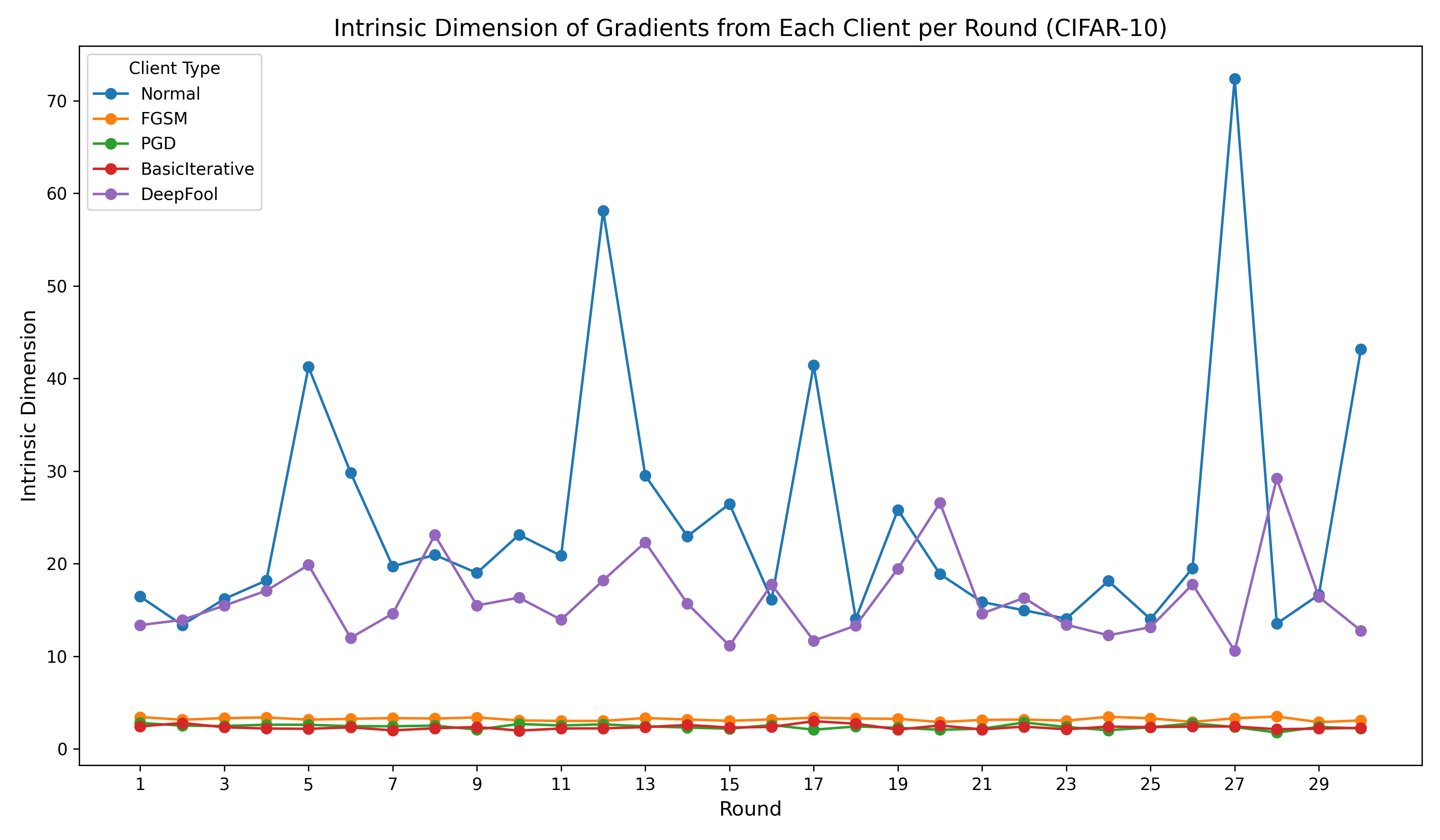}
        \caption{Raw ID Values (CIFAR-10)}
    \end{subfigure}
    \hfill
    \begin{subfigure}[b]{0.48\textwidth}
        \includegraphics[width=\linewidth]{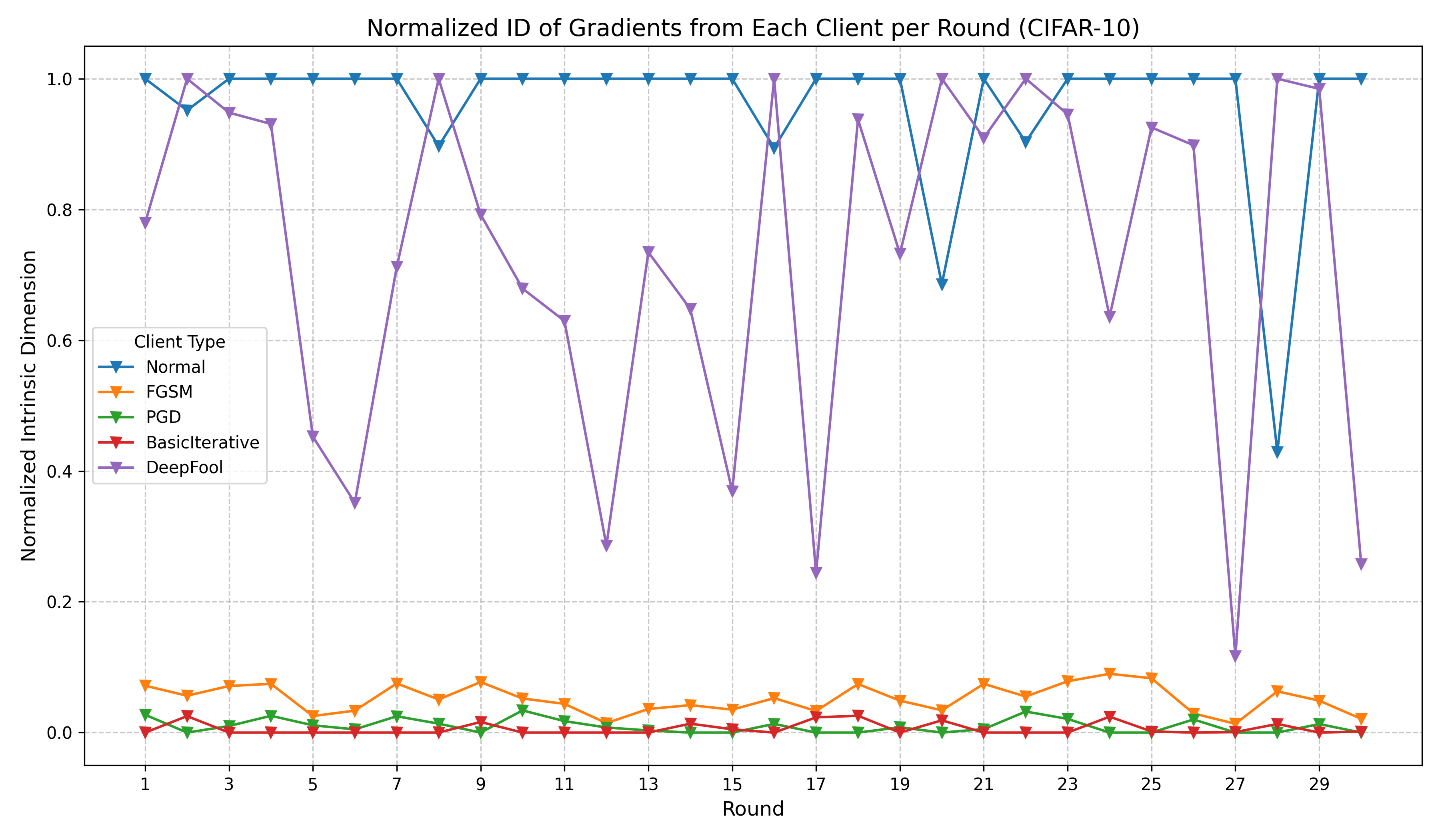}
        \caption{Normalized ID Values (CIFAR-10)}
    \end{subfigure}
    \caption{Batch-wise detection results on CIFAR-10. The trend of ID-based separability remains robust on this more complex benchmark.}
    \label{fig:appendix_cifar10_results} 
\end{figure}

Figures~\ref{fig:appendix_mnist_results} and \ref{fig:appendix_cifar10_results} present the complete batch-wise detection results for the MNIST and CIFAR-10 datasets, respectively. These visualizations supplement the SVHN results in the main paper and confirm that the clear ID-based separability between benign and malicious clients is a robust pattern that holds across all tested benchmarks.

\end{document}